\documentclass[lettersize,journal]{IEEEtran}
\usepackage{amsmath,amsfonts}
\usepackage{algorithmic}
\usepackage{algorithm}
\usepackage{array}
\usepackage{textcomp}
\usepackage{stfloats}
\usepackage{url}
\usepackage{verbatim}
\usepackage{graphicx}
\usepackage{cite}
\usepackage{multirow}
\usepackage{color}
\usepackage{subcaption}
\usepackage{stackengine}
\usepackage{booktabs}
\usepackage{colortbl}
\usepackage{xcolor}
\usepackage{ctable}
\usepackage[misc]{ifsym}
\definecolor{mygray}{RGB}{236,236,236}
\usepackage[pagebackref=false,breaklinks=true,letterpaper=true,colorlinks,linkcolor=red,bookmarks=false]{hyperref}

\newcommand{\vphan}{\vphantom{$\int^1_0$}}

\newcommand{\eg}{\textit{e.g.}}
\newcommand{\ie}{\textit{i.e.}}
\newcommand{\etal}{\textit{et al.}~}
\newcommand{{\ourmodule}}{$\mathrm{M^3Att}$}
\newcommand{\ourdecoder}{$\mathrm{M^3Dec}$}

\begin{document}

\title{Multi-Modal Mutual Attention and Iterative Interaction for Referring Image Segmentation} 

\author{Chang~Liu,
        Henghui~Ding,
        Yulun~Zhang,
        Xudong~Jiang,~\IEEEmembership{Fellow,~IEEE}~
\thanks{${\textrm{\Letter}}$ Corresponding author: Henghui Ding.}
}

\markboth{IEEE TRANSACTIONS ON IMAGE PROCESSING}%
{Shell \MakeLowercase{\textit{et al.}}: A Sample Article Using IEEEtran.cls for IEEE Journals}

\maketitle

\begin{abstract}
   We address the problem of referring image segmentation that aims to generate a mask for the object specified by a natural language expression. Many recent works utilize Transformer to extract features for the target object by aggregating the attended visual regions. However, the generic attention mechanism in Transformer only uses the language input for attention weight calculation, which does not explicitly fuse language features in its output. Thus, its output feature is dominated by vision information, which limits the model to comprehensively understand the multi-modal information, and brings uncertainty for the subsequent mask decoder to extract the output mask. 
   To address this issue, we propose Multi-Modal Mutual Attention ({\ourmodule}) and Multi-Modal Mutual Decoder ({\ourdecoder}) that better fuse information from the two input modalities. Based on {\ourdecoder}, we further propose Iterative Multi-modal Interaction ($\mathrm{IMI}$) to allow continuous and in-depth interactions between language and vision features. Furthermore, we introduce Language Feature Reconstruction ($\mathrm{LFR}$) to prevent the language information from being lost or distorted in the extracted feature.
   Extensive experiments show that our proposed approach significantly improves the baseline and outperforms state-of-the-art referring image segmentation methods on RefCOCO series datasets consistently.
\end{abstract}

\section{Introduction}

\IEEEPARstart{R}{eferring} image segmentation aims at generating mask for the object referred by a given language expression in the input image~\cite{GRES,liu2022instance,ding2020phraseclick}. Since being proposed in 2016~\cite{hu2016segmentation}, this problem has been widely discussed by many researchers, while there are still a lot of issues remaining to be addressed. 
One of the biggest challenges is that this task requires the reasoning of multiple types of information like vision and language, but the unconstrained expression of natural language and the diversity of objects in scene images bring huge uncertainty to the understanding and fusion of multi-modal features.

\begin{figure}[t]
   \begin{center}
      \includegraphics[width=0.9\linewidth]{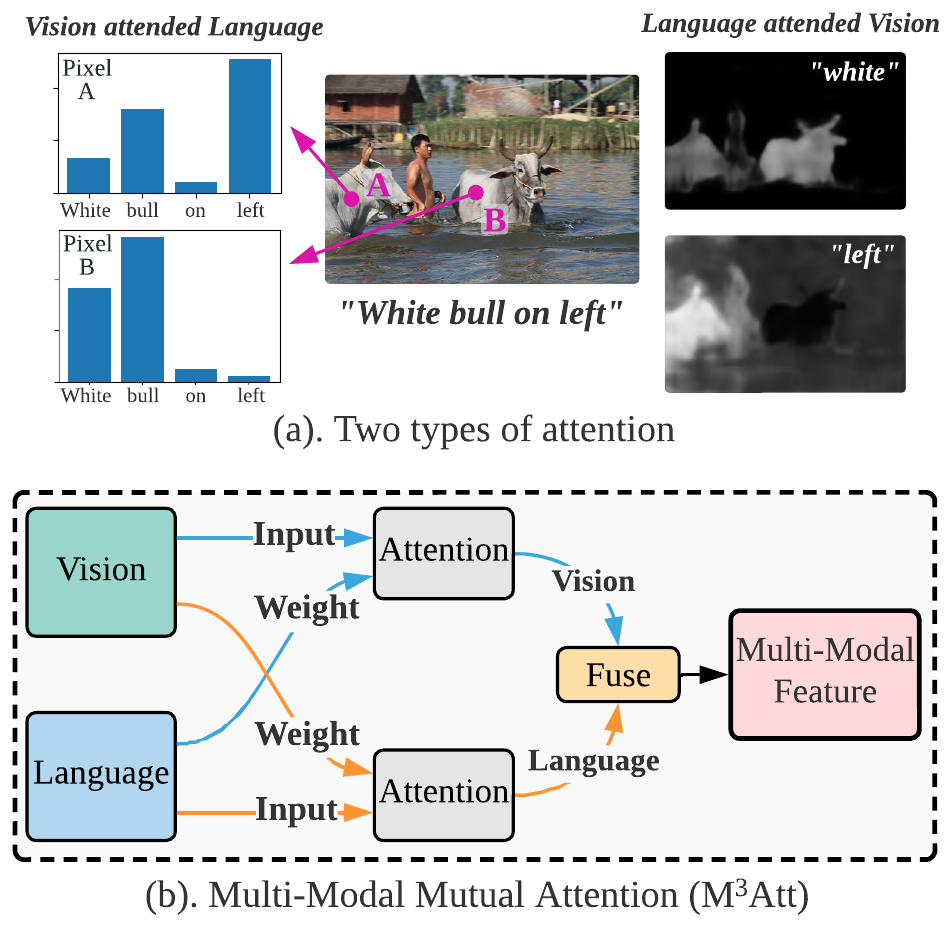}
   \end{center}
   \caption{(a). An illustration of two attention types in referring segmentation. (b). Our proposed Multi-Modal Mutual Attention ($\mathrm{M^3Att}$). (Best viewed in color)}
   \label{fig:fig1}
\end{figure}

Recently, the attention-based network has become an attractive framework for building vision models. Originally introduced for Natural Language Processing (NLP) tasks, Transformer~\cite{vaswani2017attention} is naturally suitable for solving multi-modal tasks, especially CV-NLP tasks \cite{liu2021swin,kamath2021mdetr,li2023transformer} like referring segmentation. 
Most previous works \cite{yang2021lavt,jing2021locate,dingiccv2021} utilize the generic attention mechanism to model the relationship between language and vision information.
The generic attention mechanism highlights the most relevant image region for each word in the language input, as shown in the right part of \figurename{~\ref{fig:fig1}}~\!(a). By aggregating the input vision features according to the generated attention weights, as shown in the blue path in \figurename{~\ref{fig:fig1}}~\!(b), the derived feature can describe each word using the combination of vision features. As the language feature is only used for calculating the attention weights, we call it language-attended \textit{vision} feature (LAV). 

The aforementioned attention mechanism is useful for processing vision information. However, since the referring segmentation is a multi-modal task, the language information is also essential. Thus, for processing the language information, a natural way is to introduce another type of attention that outputs language features. For each pixel in the image, we can find the words that are most relevant to it, as shown in the left part of \figurename{\ref{fig:fig1}}~\!(a). By aggregating the features of these words together according to the attention weights, a set of vision-attended \textit{language} features (VAL) for each image pixel can be derived. In contrast to LAV which is a set of vision features, VAL describes each pixel using language features. 
However, both VAL and LAV have limitations: they are both essentially single-modal features and only represent a part of the multi-modal information.
For example, VAL is a set of language features for describing pixels, but the inherent vision feature of each pixel itself is not preserved. We argue that a holistic and better understanding of multi-modal information can be get by fusing features of two modalities together. However, this is not achievable in the generic single-modal attention mechanism.

Motivated by this, we empower the generic attention mechanism with feature fusing functionality, and design a Multi-Modal Mutual Attention ($\mathrm{M^3Att}$) mechanism. It integrates two types of attention into one module, as shown in \figurename~\ref{fig:fig1}~\!(b). Our $\mathrm{M^3Att}$ has two attention pathways. One pathway (orange path) processes and outputs the vision-attended \textit{language} feature, while the other one (blue path) processes and outputs language-attended \textit{vision} feature. Two sets of features are then densely fused together, generating a real multi-modal feature with in-depth interaction of vision and language information. Using this $\mathrm{M^3Att}$ mechanism, we further design a Multi-Modal Mutual Decoder ($\mathrm{M^3Dec}$) as an optimized feature fuser and extractor for multi-modal information, which greatly enhances the performance of the model for referring segmentation.

Next, we address the modal imbalance issue in the attention-based network. Due to the characteristic of the Transformer's decoder architecture, in {\ourdecoder} as well as most attention-based works \cite{dingiccv2021,jing2021locate}, the language feature is only once inputted into the decoder at the first layer. In contrast, vision information is inputted to every decoder layer. 
This implies a modal imbalance issue: the network will tend to focus more on the vision information, and the language information may be faded away during the information propagation along the network. 
This issue will limit the strong feature fusing ability 
because of the lack of direct language input.
From this point, we propose Iterative Multi-modal Interaction (IMI), which continuously transforms the language feature and enhances the significance of language information in the multi-modal feature at each layer of the $\mathrm{M^3Dec}$, to fully leverage its fusing ability.

Furthermore,
since the ground-truth segmentation mask is the only supervision, it cannot give direct and effective feedback to encourage the model to keep the language information from being lost.
Also, as the IMI has a function of transforming the language feature,
it is helpful to protect the integrity of the language information in the multi-modal information, and prevent them from being lost and distorted. We hence propose the Language Feature Reconstruction (LFR), which protects the validity of language information in the multi-modal features in {\ourdecoder}.
A language reconstruction loss is then introduced to supervise the multi-modal features directly.

Overall, the contributions of this work can be summarized as follows:
\begin{enumerate}
   \item We propose Multi-Modal Mutual Attention (\ourmodule) and Multi-Modal Mutual Decoder (\ourdecoder) for better processing and fusing multi-modal information, and build a referring segmentation framework based on it.
   \item We propose two modules: Iterative Multi-Modal Interaction (IMI) and Language Feature Reconstruction (LFR), to further promote an in-depth multi-modal interaction in $\mathrm{M^3Dec}$.
   \item The proposed approach achieves new state-of-the-art referring image segmentation performance on RefCOCO series datasets consistently.
\end{enumerate}

\section{Related Works} \label{sec:rw}
In this section, we discuss methods that are closely related to this work, including referring segmentation and transformer.

\subsection{Referring segmentation} 
Referring segmentation is inspired by the task of referring comprehension \cite{wang2019neighbourhood,zhuang2018parallel,yang2020improving,liao2020real}. Different from semantic segmentation~\cite{ding2020semantic,shuai2018toward,ding2018context,ding2019boundary,ding2019semantic} based on pre-defined categories, referring segmentation predicts segmentation mask according to a given language expression. Defined in~\cite{hu2016segmentation}, Hu \etal introduce the classic one-stage method for referring segmentation. They firstly extract features from image and language respectively, then fuse them together and apply a FCN (Fully Convolutional Network)~\cite{long2015fully} on the fused feature. In~\cite{liu2017recurrent}, Liu \etal propose a recurrent model that utilizes word features in the sentence.~\cite{margffoy2018dynamic} and~\cite{Chen_lang2seg_2019} use the language feature to generate a set of filter kernels, then apply them on the image feature. Later, Yu \etal propose to add the word features to derive the attention weights in the later stage of the network after the tile-and-concatenate preliminary fusion is done~\cite{ye2019cross}. They design an attention module like~\cite{wang2018non,vaswani2017attention} that utilizes the word features on the multi-modal feature, achieving remarkable performance. With a similar pipeline, in~\cite{hu2020bi}, Hu \etal propose a bi-directional attention module to further utilize the features of words. In~\cite{hui2020linguistic}, Hui \etal propose to analyse the linguistic structure for better language understanding. Yang \etal\cite{yang2021bottom} use explainable reasoning. Luo \etal\cite{luo2020multi} propose a novel pipeline that merges referring segmentation and referring comprehension together, but in terms of language feature fusion, it still uses a similar multi-modal fusing technique as~\cite{ye2019cross}. Some other works propose special language usages, for example, Yu \etal\cite{yu2018mattnet} adopt a two-stage pipeline like referring comprehension methods~\cite{luo2017comprehension,hu2017modeling,hu2016natural,liu2017referring,yu2017joint,zhang2017discriminative}. Ding \etal\cite{dingiccv2021} use the language feature to generate query vectors for a transformer-based network. Feng \etal\cite{feng2021encoder} propose to utilize the language feature earlier in the encoder stage. Kamath \etal\cite{kamath2021mdetr} use transformer-based backbones\cite{devlin2018bert} for processing language inputs. Most recently, CRIS \cite{wang2022cris} proposes to use multi-modal large model CLIP~\cite{radford2021learning} to address the referring segmentation task. Yang \etal \cite{yang2021lavt} and Kim \etal \cite{kim2022restr} designed more advanced transformer architectures, and achieved impressive performance.
However, for most of the previous works, a common point is that their language information is injected into the multi-model feature at some certain ``steps''. For example, in the earlier works there is only a one-time fusion and all subsequent operations are applied on the fused features. For most recent works, the language information is used twice:
one for tile-and-concatenate preliminary fusion and the other as auxiliary information like attention module inputs. Instead, the language information in our network is iteratively utilized through the whole prediction process, establishing an in-depth interaction between features from two modalities.
Besides, most previous networks are unaware of whether the language information is lost during the propagation of the network. Our network ensures that the language information is kept until the rear stage of the network, promoting it to fully interact with information from the other modality.

\subsection{Transformer} 
Transformer is firstly introduced by Vaswani \etal~for Natural Language Processing (NLP) task \cite{vaswani2017attention}. Quickly it becomes a popular sequence-to-sequence model in the NLP area. Thanks to its strong global relationship modeling ability, it was migrated into the Computer Vision (CV) area recently and has achieved good performance in many tasks, such as image classification~\cite{dosovitskiy2020image}, deblurring~\cite{lin2022flow}, object detection~\cite{carion2020end}, semantic segmentation~\cite{xie2021segformer,PADing}, instance segmentation~\cite{cheng2021per,D2Zero}, and video segmentation~\cite{MOSE,ke2022video,sun2022coarse}. Its good performance in various areas also suggests its potential in handling multi-modal information~\cite{liu2021swin}, and there are several works on multi-modal transformers. For example, Radford \etal design a network that uses the natural language to supervise a vision model~\cite{radford2021learning}. Kim~\etal~propose a large scale pretrained model for vision-language tasks~\cite{kim2021vilt}. However, most of the relevant works are built upon the generic transformers that are originally designed for a single modality, \eg, language \cite{vaswani2017attention} or vision \cite{liu2021swin}. These methods are not optimized for processing multi-modal information, so they lack some functions for multi-modal features, for example, feature fusion. In this work, we propose a Mutual Attention mechanism, which is designed for multi-modal features. It accepts inputs from multiple modalities, enables them to interact with each other, and densely fuses them together, so as to output a true multi-modal feature.

\begin{figure*}[t]
   \begin{center}
      \includegraphics[width=0.996\linewidth]{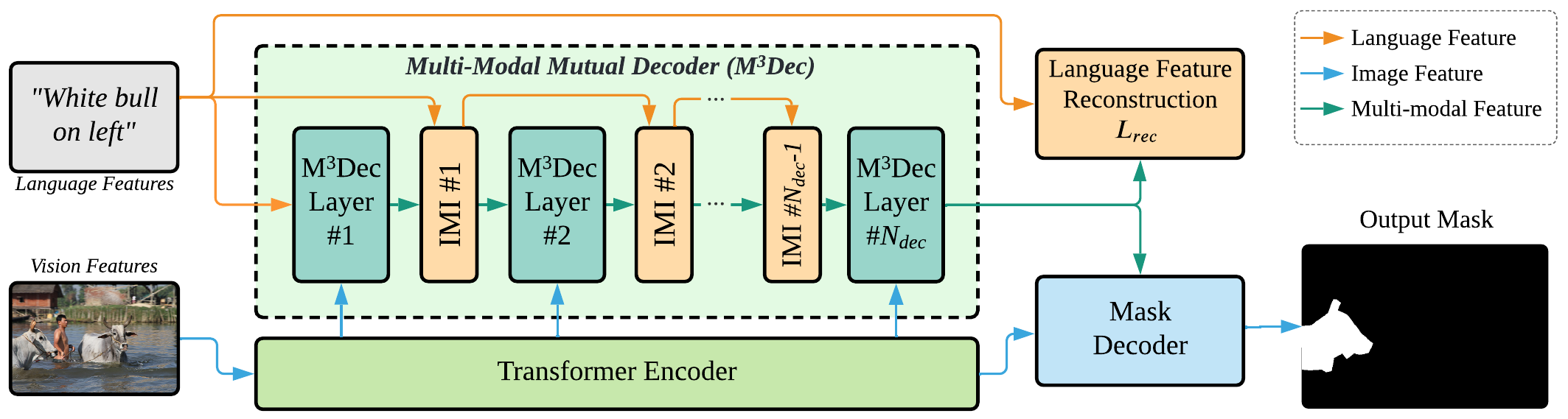}
   \end{center}
   \vspace{-3mm}
   \caption{(Best viewed in color) The overall architecture of the proposed approach. We propose Multi-Modal Mutual Decoder ({\ourdecoder}) to fuse and process the multi-modal information from two inputs.}
   \label{fig:netarch}
\end{figure*}

\section{Methodology}

The overview architecture of our proposed approach is shown in \figurename~\ref{fig:netarch}. The network's inputs include an image $I$, and a language expression $T$ containing $N_t$ words. Following previous works~\cite{hu2016segmentation,luo2020multi,dingiccv2021}, we first extract two sets of input backbone features: image feature $F_{vis}$ from $I$ using a CNN backbone, and language feature $F_t$ and $F_t'$ from $T$ using a bi-directional LSTM. The image feature, $F_{vis}$ has the shape of $H\times W\times C$, where $H$ and $W$ denote height and width respectively, $C$ is the number of channels. For the language feature, the hidden states of the LSTM $F_t\in R^{N_t\times C}$ represent the feature for each word, while the final state output $F_t'$ is used as the representation of the whole sentence. The channel number of language features is also $C$ for the ease of fusion.

Then we send vision feature to a transformer encoder with $N_{enc}$ layers to obtain deep vision information $F_{enc}$. Next, we input $F_{enc}$ into our proposed Multi-Modal Mutual Decoder ({\ourdecoder}) and Iterative Multi-modal Interaction (IMI), which give an in-depth interaction for the multi-modal information. Finally, the Mask Decoder takes the output from both transformer encoder and {\ourdecoder}, and generates the output mask. Moreover, we propose a Language Feature Reconstruction (LFR) module to encourage language usage in the {\ourdecoder}, and prevent that the language information from being lost at the rear layers of the network.
The details of each part will be introduced in the following sections.

\subsection{Multi-Modal Mutual Attention} \label{sec:m3att}

\begin{figure*}[t]
   \begin{center}
      \includegraphics[width=\linewidth]{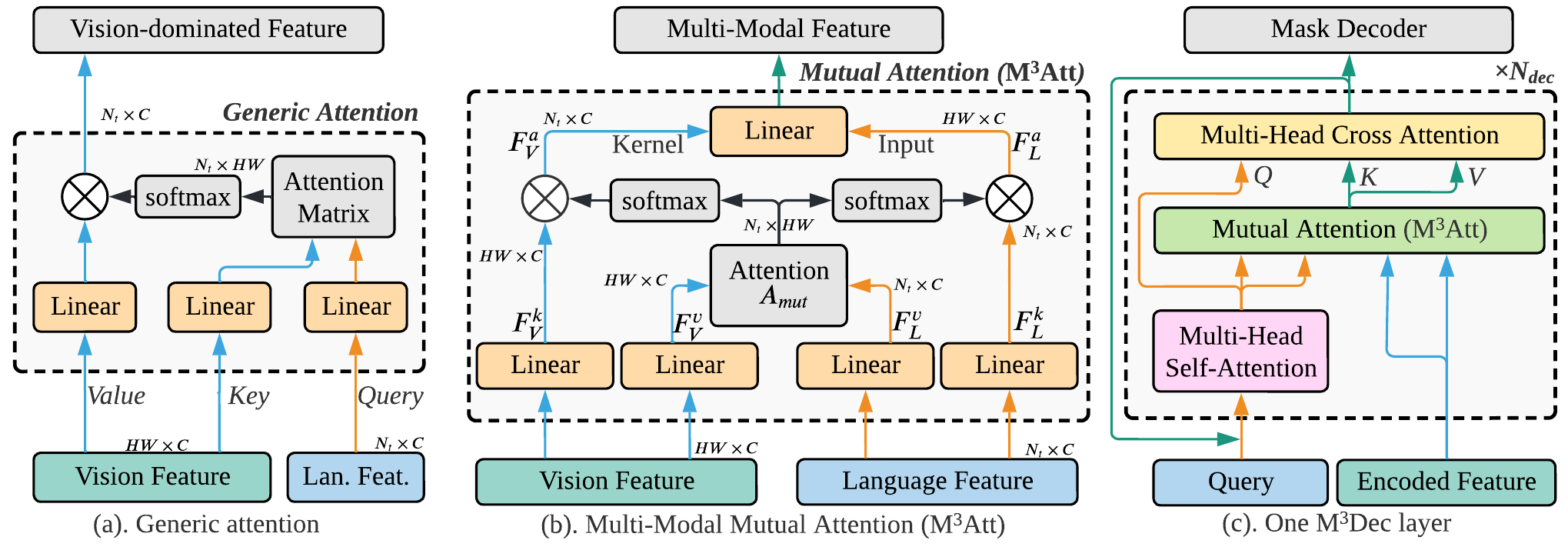}
   \end{center}
   \vspace{-3mm}
   \caption{The architecture of: (a). the generic attention mechanism; (b). the proposed Multi-Modal Mutual Attention ({\ourmodule}); (c). one layer of the proposed Multi-Modal Mutual Decoder ({\ourdecoder}).}
   \label{fig:symatt1}
\end{figure*}

As mentioned above, most previous works use the generic attention mechanism for processing multi-modal information. \figurename~\ref{fig:symatt1}~\!(a) gives an example of such kind of mechanism, similar to \cite{dingiccv2021}. Features from two modalities (query and key) are used to derive an attention matrix, that is then used to aggregate the vision feature for each word. In this process, the language feature is only used to generate the attention weights that indicate the significances of regions in the vision feature. Hence, language information is not directly involved in the output so that the output can be viewed as a reorganized single-modal vision feature. Even worse, this single-modal vision output is used alone as a query in the successive transformer decoder, dominating information in decoder.
As a result, language information will be dramatically lost in the decoder.
Thus we argue that the generic attention mechanism is good for \textit{processing} features from the value input, but it lacks the ability of \textit{fusing} features from two modalities. 
So, if it is used to process multi-modal information, the query input is not fully utilized, and features of two modalities are not densely fused and interacted.

To address this issue, we propose a Multi-Modal Mutual Attention ({\ourmodule}), as shown in \figurename{~\ref{fig:symatt1}}~\!(b). {\ourmodule} takes inputs from two modalities, and transforms each of them into two roles: key and value. This enables us to equally treat features from both modalities and fuse them together. Herein we use language feature $F_t\in R^{N_t\times C}$ and vision feature from the output of the transformer encoder $F_{enc}\in R^{HW\times C}$ as inputs to illustrate the architecture of {\ourmodule}. Firstly, we use linear layers to project the language features into keys $F^k_L$ and values $F^v_L$, and similarly project the vision features into $F^k_V$ and $F^v_V$. Next, we use the two keys from two modals to generate the mutual attention matrix:
\begin{equation}
   A_{mut} = \frac{1}{\sqrt{C}} F^k_L (F^k_V)^T, 
\end{equation}
which is a multi-modal attention matrix with shape of $N_t\times HW$, describing the relationship strength from all elements of one modal to all elements of the other modal. $\frac{1}{\sqrt{C}}$ is the scaling factor \cite{vaswani2017attention}. Then unlike the generic attention that only applies the attention matrix on one modal, we normalize the mutual attention matrix in both axes and apply it on features from the both modals:
\begin{subequations}
   \begin{align}
      F_V^a &= \text{softmax}(A_{mut}) F^v_V,  \label{eq:Asoftmaxv} \\
      F_L^a &= \text{softmax}({A_{mut}}^T) F^v_L.  \label{eq:Asoftmaxl}
   \end{align}
\end{subequations}

\textbf{Language-attended \textit{vision} feature (LAV), \boldmath{$F_V^a$}}: Softmax normalization is applied along each $HW\times 1$ axis of the mutual attention matrix $A_{mut}$, as in Eq.~(\ref{eq:Asoftmaxv}), which is then applied on the vision feature $F_V^v$ to get the language-attended \textit{vision} feature $F_V^a\in R^{N_t\times C}$. There are $N_t$ feature vectors in $F_V^a$, where each vector represents one attended vision feature corresponding to one element (word) in the language input. In other words, each vector is the vision feature weighted by a word based on its interpretation to the image. It is similar to the output of the generic attention mechanism. As the language features only participate in the attention matrix, the output is essentially still a single-modal feature. 

\textbf{Vision-attended \textit{language} feature (VAL), \boldmath{$F_L^a$}}. Another softmax normalization is applied on the transposed mutual attention  matrix, ${A_{mut}}^T$, along the $N_t\times 1$ axis, as in Eq.~(\ref{eq:Asoftmaxl}). By applying the attention matrix on the language feature $F_L^v$, we get the vision-attended \textit{language} feature $F_L^a\in R^{HW\times C}$. $F_L^a$ contains $HW$ feature vectors, where each vector represents one attended language feature corresponding to one pixel in the vision feature. In other words, $F_L^a$ is a spatial-dynamic language feature, each vector of $F_L^a$ is the language feature weighted based on a pixel's interpretation of the sentence.

\textbf{Fusing of multi-modal attended features. }Next, we use both attended vision and language features to generate the output. We treat each of the attended vision features in $F_V^a$ as a dynamic kernel of a linear layer applied on $F_L^a$: $F_{mul} = F_V^a (F_L^a)^T$.
Thus, the result is a true multi-modal feature $F_{mul}\in R^{N_t\times HW}$, where $N_t$ is the sequence length of query and $HW$ is the channel number. Finally, a linear layer is used to project the channel number back to $C$, and generates the output of this {\ourmodule} module. Notably, the mutual attention matrix $A_{mul}$ for two softmax functions in Eq.~(\ref{eq:Asoftmaxv}) and Eq.~(\ref{eq:Asoftmaxl}) is default to be shared, but it can also be independently computed. More details will be discussed in the experiments. It is also worth noting that our {\ourmodule} is not limited to deal with language and vision features but can accept and fuse any two modalities.

Based on {\ourmodule}, we build a Multi-Modal Mutual Decoder ({\ourdecoder}). {\ourdecoder} has $N_{dec}$ stacked layers
, as shown in \figurename{~\ref{fig:symatt1}}~\!(c) for one layer. Each layer has the same architecture and takes two inputs: encoded feature and query. Here we use the first {\ourdecoder} layer in our network to illustrate the layer architecture, in which the language feature $F_t$ is taken as the query input, and transformer encoder output $F_{enc}$ is taken as the encoded feature. Inside the layer, firstly a multi-head self-attention layer is applied on the query input, outputting a set of query features $F_q$. Next, a {\ourmodule} module is used to fuse two sets of features: one is the query feature that is derived from the language feature and other is the transformer encoder output that has rich vision clues. The resulting multi-modal feature is further queried again by the query feature using Multi-Head Cross Attention, generating the output of this decoder layer. In this step, we use the multi-modal feature as value input, so that the output can keep its property as a multi-modal feature. The output of each {\ourmodule} layer is used as the query input to its successive layer, replacing the language feature of the first layer. The output of the final layer is sent to the Mask Decoder to generate the output mask.

\subsection{Iterative Multi-Modal Interaction}
\begin{figure}[t]
   \centering
      \includegraphics[width=\linewidth]{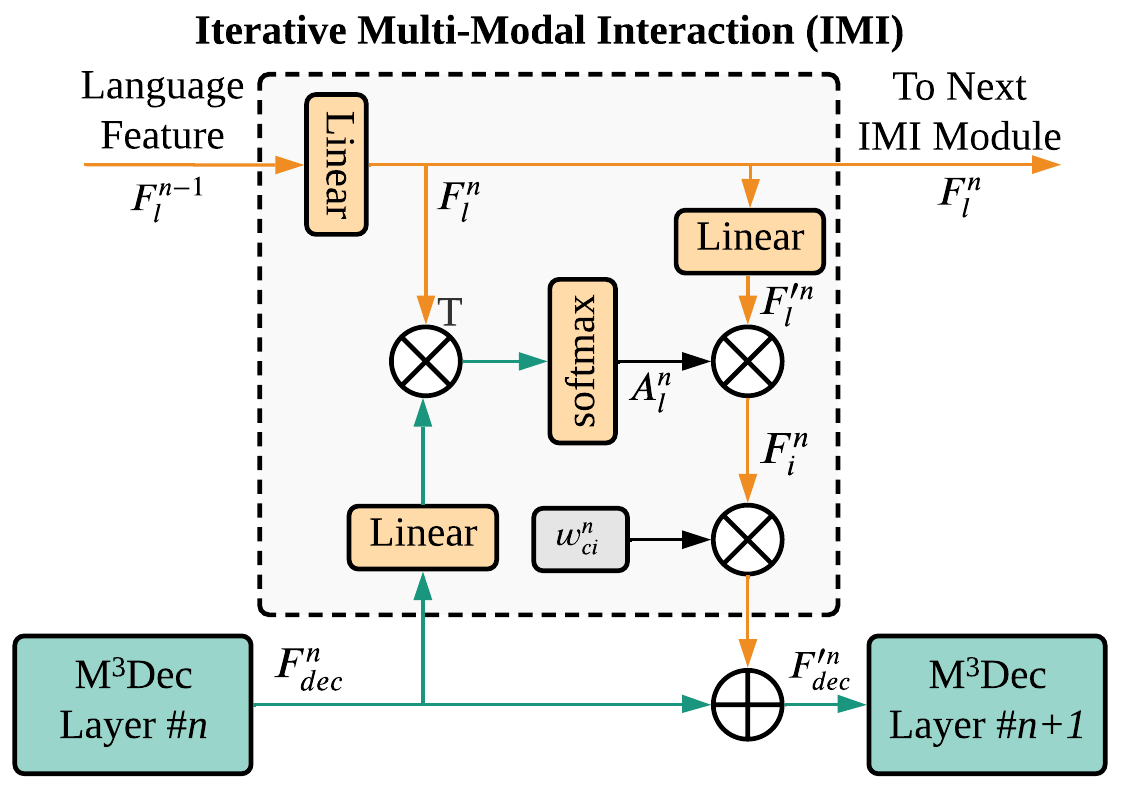}
      \caption{The architecture of one block of the Iterative Multi-Modal Interaction (IMI) module.}
      \label{fig:ci}
\end{figure}
Due to the characteristic of the attention-based network, as discussed above, in {\ourdecoder}, the output of the previous layer is used as the query input to the next layer. Thus, from the second layer onwards, the layer input will be the encoded feature $F_{enc}$ and the output of its previous layer. In other words, vision information $F_{enc}$ is directly inputted into every layer since the beginning, but in contrast, the language feature is only inputted once at the first decoder layer, as shown in \figurename{~\ref{fig:netarch}}.
This leads to a modal imbalance issue, and may cause the language information to be faded away in the rear stage of the network. This issue also exists in many previous transformer-based works~\cite{dingiccv2021,jing2021locate}. Although {\ourmodule} addresses this issue by fusing the language and vision information in the first layer using its strong multi-modal fusing ability, without language information inputted in the later stages, its feature fusing potential is not fully leveraged. From this point, we propose to inject the language information into {\ourdecoder} at every layer.

Besides, as features are propagated to higher layers, the model's understanding of language information becomes deeper. This also causes that different layers will focus on different types of information. For example, features from lower layers do not have a contextual understanding of the relationship between language and image so that they desire more specific clues, while features from higher layers need more holistic information as they already have a better understanding of image and language. Therefore, it is desired to transform the language features along with the processing of the multi-modal feature.

Combining the above two points, we propose an Iterative Multi-Modal Interaction (IMI) module, which provides an opportunity for multi-modal features at different layers to query about their desired language information, and continuously inject them into the decoder. The IMI blocks are inserted between each two successive {\ourdecoder} layers. 
For the $n\textsuperscript{th}$ IMI block, as shown in \figurename~\ref{fig:ci}, it takes two inputs: the output of the $n\textsuperscript{th}$ {\ourdecoder} layer $F_{dec}^n\in R^{N_t\times C}$, and the output of the previous IMI layer $F_{l}^{n-1}\in R^{N_t\times C}$. $F_{l}^{n-1}$ is firstly transformed with a linear layer, generating the language feature of the current layer, $F_{l}^{n}$. The language input for the first IMI block is the word feature $F_{t}$. With each IMI block connected to the previous one, we create a dedicated pathway for processing the language information, parallel with the process of multi-modal information.

Next, we project the multi-modal feature using a linear layer, and compute an attention matrix for reorganizing the language features:
\begin{equation}
 A_l^n = \text{softmax}(\text{ReLU}[F_{dec}^{n} W_{a}^{n}] (F_l^n)^T),
\end{equation}
The attention matrix $A_l^n$ is then used to reform a new language feature: $F_i^n=A_l^n F_l^{\prime n}$, where $F_l^{\prime n}=\text{ReLU}[F_{l}^{n} W_{l}^{\prime n}]$, as shown in \figurename~\ref{fig:ci}. The resulting feature $F_i^n$ is then injected back into the {\ourdecoder} layer output $F_{dec}^n$ under the control of a learnable scalar $w_{ci}^n$, \ie, $ F_{dec}^{\prime n} = \text{BN}(F_{dec}^{n} + w_{ci}^n F_i^n),$
where $\text{BN}$ denotes batch normalization. Using the learnable weight allows the network to determine how much information is needed by itself, and also makes the language feature more adaptable to the multi-modal feature. The output $F_{dec}^{\prime n}$ is sent to the next {\ourdecoder} layer as the query input.

\subsection{Language Feature Reconstruction}

\begin{figure}[t]
   \begin{minipage}{0.48\textwidth}
      \includegraphics[width=\linewidth]{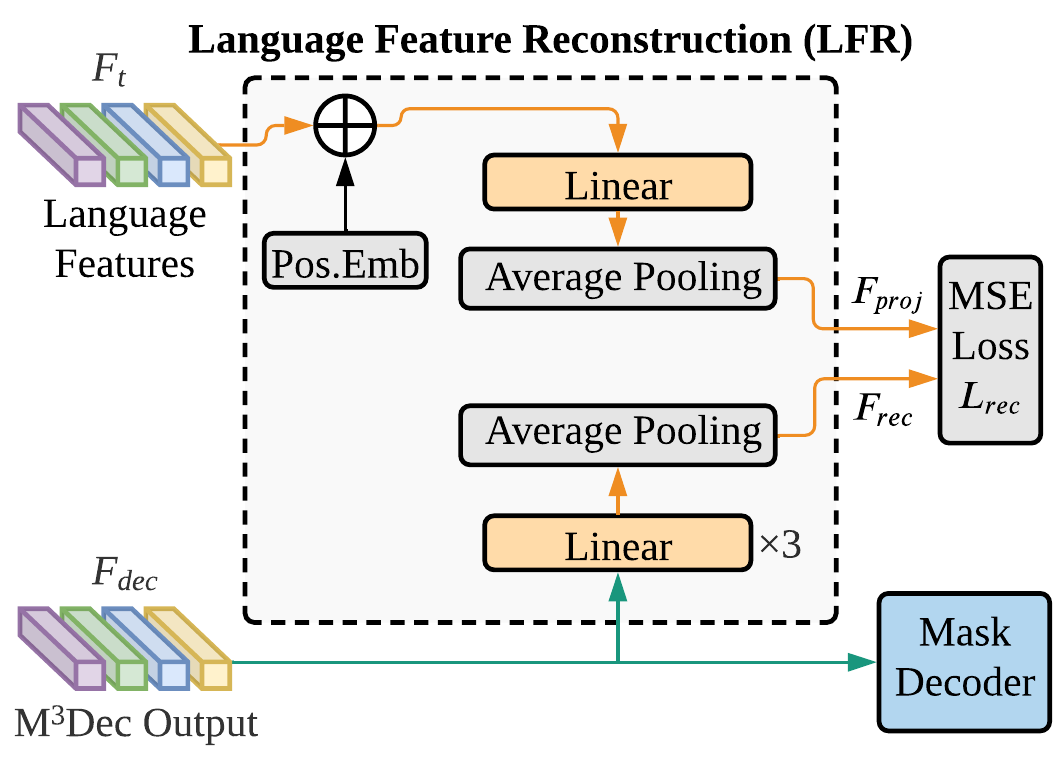}
      \caption{The Language Feature Reconstruction (LFR) module. Pos.Emb: Positional Embedding.}
      \label{fig:langcons}
   \end{minipage}%
\end{figure}
In most referring segmentation methods, the network is only supervised by the output mask loss. This implies a hypothesis: as long as the output mask matches the target object, we consider that the model has successfully understood the language information. However, this is not always true in real-world scenarios. For example, it is assumed in most datasets that there is always one and only one object in the ground-truth segmentation mask for each training sample. The network can easily learn such kind of data bias and always output one object. Therefore, for some training samples, if the network happens to ``guess'' the correct target even if the language information has been lost during the propagation, these training samples may not properly contribute to the training of the network, or even be harmful to the network to generalize.

To encourage the network to be better generalized in learning from samples and improve its resistance to the language information lost, we propose a Language Feature Reconstruction (LFR) module, located at the end of the last {\ourdecoder} layer in \figurename~\ref{fig:netarch}. The proposed LFR module tries to reconstruct the language feature from the last {\ourdecoder} output. So it ensures that the language information is well preserved through the whole multi-modal feature processing procedure. The architecture is shown in \figurename~\ref{fig:langcons}. It takes language features $F_t\in R^{N_t\times C}$, $F_t'\in R^{1\times C}$, and the output of the last {\ourdecoder} layer $F_{dec}\in R^{N_t\times C}$ as inputs. The language features $F_t$, $F_t'$ and the multi-modal feature $F_{dec}$ are projected into the same feature space for comparison.
\begin{equation}
   F_{proj} = \frac{1}{N_t+1}\sum{\text{ReLU}\Big([(F_{t} + e)\ \text{\large\copyright}\  F_t'] W_{proj}\Big)},
\end{equation}
where \begin{footnotesize}$\copyright$\end{footnotesize} is concatenation. The both \begin{footnotesize}$\copyright$\end{footnotesize} and \begin{footnotesize}$\sum$\end{footnotesize} are conducted along the  sequence length dimension (\ie, $[(F_{t} + e)\ \text{\footnotesize\copyright}\  F_t'] \in R^{(N_t+1)\times C}$).~$e$ denotes the cosine positional embedding, which adds information about the order of words in the sentence.~$W_{proj}\in R^{C\times C}$ is learnable parameters for projection, and $N_t$ is the length of the sentence for normalization.

Next, language information is reconstructed from the final multi-modal feature. As shown in \figurename~\ref{fig:langcons}, we first apply three stacked linear layers on the {\ourdecoder} output $F_{dec}$, then use an average pooling layer to shrink the sequence length dimension, producing the reconstructed language feature $F_{rec}$. The Language Feature Reconstruction loss is derived by minimizing the distance between the reconstructed language feature $F_{rec}$ that is comparable with $F_{proj}$ and the project language feature $F_{proj}$ using the Mean Squared Error Loss.

\subsection{Mask Decoder and Loss Function}

\begin{figure}[t]
   \centering
   \begin{minipage}{0.48\textwidth}
      \includegraphics[width=\linewidth]{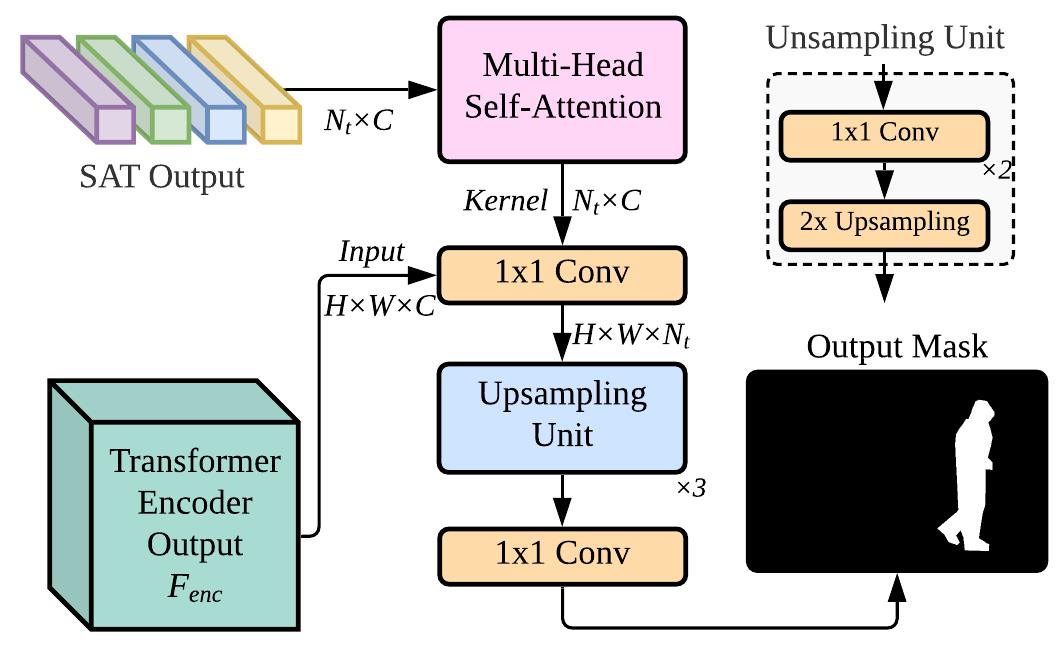}
      \caption{The Mask Decoder takes the output of the Mutual Attention Decoder ({\ourdecoder}) and the output of the Transformer encoder to form the output mask.}
      \label{fig:md}
   \end{minipage}%
\end{figure}

The last step of our framework is to extract the output mask from the multi-modal features.
In our framework, since we have a dense multi-modal interaction in the {\ourdecoder}, we would like the decoder to focus more on understanding the semantic clues in the inputs, and use a more vision-dominated feature to focus on the fine-grained vision details. Therefore, we choose to use both encoder output $F_{enc}\in R^{H\times W\times C}$ and {\ourdecoder} output $F_{dec}\in R^{N_t\times C}$ to generate the segmentation mask as shown in \figurename~{\ref{fig:md}}. The Mask Decoder firstly processes $F_{dec}$ with a self attention module. Next, the processed decoder feature serves as kernels of a $1\times 1$ convolutional layer. With $F_{enc}$ as input, $N_t$ feature maps are generated by this convolutional layer.  Finally, we use four stacked convolutional layers to output the prediction mask. Upsampling layers are inserted between convolution layers for recovers the spatial size of the mask.

The output mask is supervised by the Binary Cross Entropy Loss. The final loss function is defined as:
\begin{equation}
   \mathcal{L}=w_{mask}\mathcal{L}_{mask} + w_{rec}\mathcal{L}_{rec},
\end{equation}
where $w_{mask}$ is the weight for the mask loss $\mathcal{L}_{mask}$ and $w_{rec}$ is the weight for the Language Feature Reconstruction loss $\mathcal{L}_{rec}$.
The proposed LFR does not directly participate in the mask prediction and is computationally free during inference. It can work as a plug-in module to any existing referring segmentation methods.
\section{Experiments}

In this section, we report the experimental results of our method in comparison with previous state-of-the-art methods, and the ablation studies that verify the effectiveness of our proposed modules. We evaluate the performance by two commonly used metrics: the IoU score measures the rate of Intersection over the Union between the model's output mask and the ground-truth mask, and
the Precision@X score built on IoU. Given a threshold X, the Precision@X score computes the percentage of successful predictions that have IoU scores higher than X.

\subsection{Implementation Details}

We train and evaluate the proposed approach on three commonly-used referring image segmentation datasets: RefCOCO~\cite{yu2016modeling}, RefCOCO+~\cite{yu2016modeling}, and RefCOCOg~\cite{mao2016generation}. 
Following previous works~\cite{luo2020multi,dingiccv2021,jing2021locate}, the image features are extracted by a Darknet-53 backbone~\cite{yolo} pretrained on MSCOCO~\cite{lin2014microsoft} dataset and language embeddings are generated by GloVE~\cite{pennington2014glove}. Language expressions are padded to 15 words for RefCOCO/RefCOCO+ and 20 words for RefCOCOg. Images from the validation set and test set of the referring segmentation datasets are excluded when training the backbone. Images are resized to 416$\times$416 for CNN backbone following \cite{dingiccv2021,luo2020multi,yu2018mattnet} and 480$\times$480 for Transformer backbone following \cite{yang2021lavt,VLTPAMI}.  Channel number $C$ is fixed to 256 for the transformers and 512 for the mask decoder. The network has 2 encoder layers. The head number is 8 for all transformer layers. The weight for mask loss $w_{mask}$ is set to 1 and the Language Feature Reconstruction loss $w_{rec}$ is set to $0.1$. 
All linear layers and convolutions layers are followed by a Batch Normalization and ReLU function unless otherwise noticed. The network is trained for 50 epochs with the batch size set to 48, using the Adam~\cite{kingma2014adam} optimizer. The learning rate is set to 0.005 with a step decay schedule. We use 4 NVIDIA V100 GPUs for training and testing.
\subsection{Ablation Study}

\begin{table*}[t]
\renewcommand\arraystretch{1.1} 
   \centering
   \small
   \caption{Ablation results of number of layers of {\ourdecoder} in different settings on the validation set of RefCOCO.}\vspace{-2mm}
   \setlength{\tabcolsep}{6mm}{\begin{tabular}{r|cc|cc|cc}
         \specialrule{.1em}{.05em}{.05em}
       {\multirow{2}[0]{*}{Layers}} & \multicolumn{2}{c|}{Shared} & \multicolumn{2}{c|}{Independent} & \multicolumn{2}{c}{Generic (LAV only)} \\
       \cline{2-7}
             & {IoU} & {Pr@0.5} & {IoU} & {Pr@0.5} & {IoU} & {Pr@0.5} \\
             \hline \hline
             1     & 60.57 & 72.47 & 62.36 & 74.00 & 55.42 & 70.02 \\
             2     & 66.30 & 78.01 & 66.32 & 78.55 & 64.12 & 75.33 \\
             3     & \textbf{67.88} & \textbf{79.01} & 67.80 & 78.94 & 64.40 & 75.81 \\
             4     & 67.82 & 78.95 & 67.76 & 78.99 & 64.42 & 75.79 \\
       \specialrule{.1em}{.05em}{.05em}
       \end{tabular}}%
     \label{tab:SADablation}%
\end{table*}%

\begin{table}[t]
\renewcommand\arraystretch{1.1} 
   \centering
   \small
   \caption{Ablation study of components on the validation set of RefCOCO.}\vspace{-2mm}
   \setlength{\tabcolsep}{3.6mm}{\begin{tabular}{r|l|cc}
         \specialrule{.1em}{.05em}{.05em}
         No. & Model                     & IoU            & Pr@0.5         \\
         \hline \hline
         \#0 & Baseline (Generic LAV)    & 62.04          & 73.52          \\
         \#1 & Baseline (\ourmodule)     & 65.56          & 76.66          \\
         \#2 & Baseline + IMI            & 66.70          & 77.62          \\
         \#3 & Baseline + IMI$^*$        & 65.92          & 76.80          \\
         \hline
         \#4 & Ours      & \textbf{67.88} & \textbf{79.01} \\
         \specialrule{.1em}{.05em}{.05em}
      \end{tabular}}%
   \label{tab:ablation}%
\end{table}%

\begin{table}[t]
\renewcommand\arraystretch{1.1} 
   \centering
   \small
   \caption{Ablation study of settings of the Mask Decoder on the validation set of RefCOCO.}\vspace{-2mm}
   \setlength{\tabcolsep}{4.6mm}{\begin{tabular}{l|ccc}
         \specialrule{.1em}{.05em}{.05em}
         Model                  & {IoU}          & {Pr@0.5}       & {Pr@0.9}       \\
         \hline \hline
         VLT\cite{dingiccv2021} & 66.05          & 78.64          & 13.81          \\
         Concate                & 66.58          & 78.90          & 15.55          \\
         Ours                   & \textbf{67.88} & \textbf{79.01} & \textbf{17.70} \\
         \specialrule{.1em}{.05em}{.05em}
      \end{tabular}}%
   \label{tab:MDablation}%
\end{table}%
 
We do several ablation experiments to show the effectiveness of each proposed module in our framework. The results are reported in \tablename~\ref{tab:SADablation} and \tablename~\ref{tab:ablation}. 

\textbf{\boldmath{\ourmodule}, VAL and LAV.} As mentioned in Section \ref{sec:m3att}, the attention matrix $A_{mut}$ for two attended features in the {\ourmodule} can be computed in two ways: shared or independent. We report the results of the two {\ourmodule} settings over the generic attention mechanism in \tablename~\ref{tab:SADablation}. For the shared setup, two $A_{mul}$ in Eq.~(\ref{eq:Asoftmaxl}) and Eq.~(\ref{eq:Asoftmaxv}) are identical. For the independent setup, the attention module has two extra linear project layers applied on two inputs, generating two $A_{mul}$, one for LAV and the other for VAL. As shown in \tablename~\ref{tab:SADablation}, when the layer numbers are lower, the independent setup performs better than the shared setup. However, with the increase of the layer numbers, the performance of two settings gradually gets similar. When there are 3 decoder layers, the shared setup even slightly outperforms the independent setup. We presume that this is because the independent setup has extra parameters, thus there is a performance gap when the layer numbers are smaller and the parameter numbers are not enough. We use the Shared {\ourmodule} with 3 decoders as the default setup of our network.

Besides, to prove the importance of the feature fusing ability of our transformer, we compare the performance of \ourmodule with the generic transformer, which can only generate single-modal features. Firstly we test the generic-attention base transformer that only use the LAV feature, \ie, word features serve as the query input and vision features serve as key and value input, similar as the transformer architecture in VLT~\cite{dingiccv2021}. The results are reported in the ``Generic (LAV)'' column. It can be seen that our module greatly enhances the performance, showing that multi-modal features are essential for understanding the vision and language inputs. Finally, because VAL feature are single modal language feature that are not feasible for generating masks alone, the transformer with only VAL features, \ie, using language features as key/value input and using vision feature as query, fails to converge. Above two experiments show that VAL feature is a great assistance to the LAV feature.

\textbf{IMI and LFR.} The ablation results of IMI and LFR are reported in \tablename~\ref{tab:ablation}. In the baseline model, both IMI and LFR are removed. In Model \#1, we validate the effectiveness of the IMI. It brings a performance gain of $1.14\%$ in terms of IoU and $0.96\%$ in terms of Pr@0.5. Besides in Model \#2, to verify our motivation that different layers in the {\ourdecoder} need different language information, we simplify the transforming function of the IMI by replacing the $F_i^n$ in \figurename{~\ref{fig:ci}} with the language feature $F_t$, making that all {\ourdecoder} layers receive the same language feature. This method only gives a very slight performance improvement of $0.36\%$ IoU over baseline, showing that by constructing the transformation pathway for language information, the IMI successfully extracts appropriate information for different feature processing stages. Finally, we add the Language Feature Reconstruction (LFR) module. Compared with Model \#1, it brings an improvement on the IoU by $1.18\%$. Totally, the IMI and LFR bring over $2\%$ improvement in terms of the both IoU and Pr@0.5.

\textbf{Mask Decoder.} In \tablename~\ref{tab:MDablation}, we report the performance of our Mask Decoder against other variants. In the first model, we use the Mask Decoder from VLT~\cite{dingiccv2021}, which utilizes only the output of the decoder. In the second model, rather than using the {\ourdecoder} output as the convolution kernel, we sum and concatenate them with the transformer encoder output. By comparing the precision metrics, our Mask Decoder increases the Pr@0.9 metric by $3.89\%$ from the baseline model and $2.15\%$ from the concatenating method, showing that both the encoder and decoder information are essential to the performance, and our mask decoder can better preserve the fine-grained image details while not losing the targeting ability.

\begin{figure*}[t]
   \begin{center}
      \begin{subfigure}[t]{0.5\linewidth}
         \scriptsize
         \stackon{\includegraphics[width=0.325\linewidth]{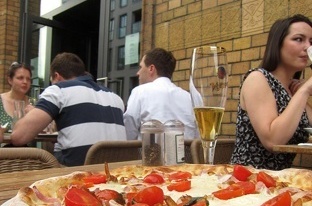}}{Image\vphan}\hfill%
         \stackon{\includegraphics[width=0.325\linewidth]{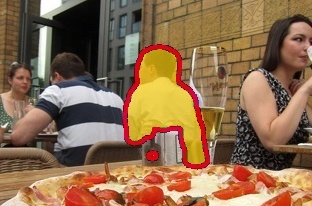}}{Baseline\vphan}\hfill%
         \stackon{\includegraphics[width=0.325\linewidth]{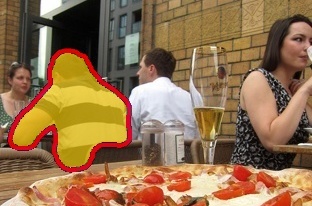}}{Ours\vphan}
         \vspace{-0.04in}
         \caption{\centering \scriptsize{man with his back away from us in a blue and white striped shirt eating}}
      \end{subfigure}%
      \begin{subfigure}[t]{0.5\linewidth}
         \scriptsize
         \stackon{\includegraphics[width=0.325\linewidth]{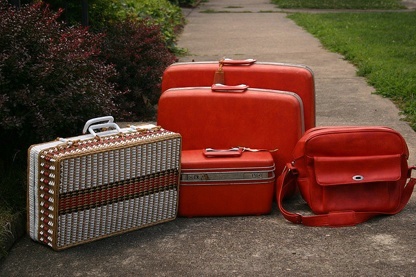}}{Image\vphan}\hfill%
         \stackon{\includegraphics[width=0.325\linewidth]{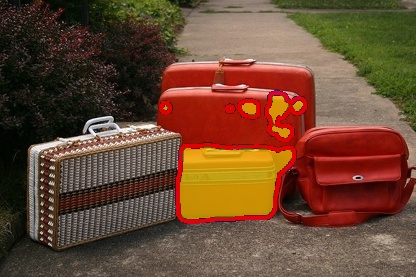}}{Baseline\vphan}\hfill%
         \stackon{\includegraphics[width=0.325\linewidth]{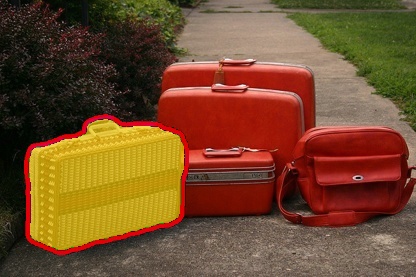}}{Ours\vphan}
         \caption{\centering \scriptsize{the suitcase that isn't red}}
      \end{subfigure}%
      \caption{(Best viewed in color) Qualitative comparison with the baseline model. The proposed approach is able to solve the hard cases that cannot be handled by the baseline model.}
      \label{fig:ablation}
   \end{center}
\end{figure*}
\begin{figure*}[t]
   \begin{center}
      \begin{subfigure}[t]{0.49\linewidth}
         \scriptsize
         \stackon{\includegraphics[width=0.325\linewidth]{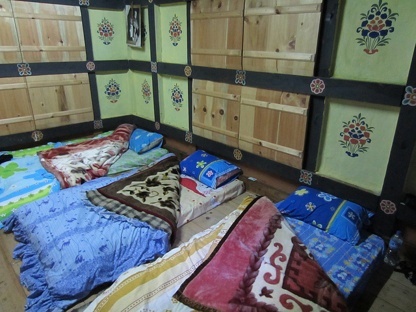}}{Image\vphan}\hfill%
         \stackon{\includegraphics[width=0.325\linewidth]{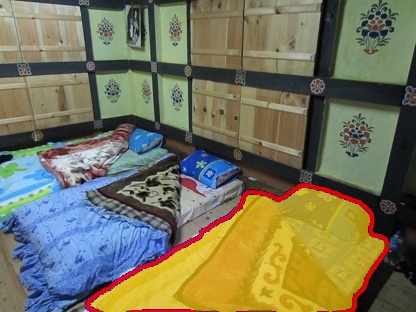}}{Prediction\vphan}\hfill%
         \stackon{\includegraphics[width=0.325\linewidth]{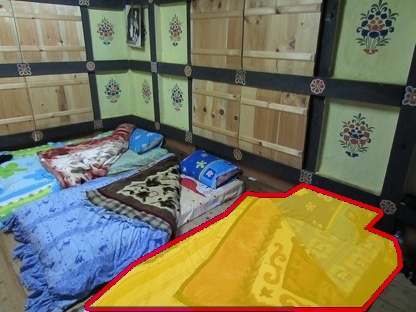}}{Ground-Truth\vphan}
         \vspace{-0.04in}
         \caption{\centering \scriptsize{matress pink and yellow in color and on the blue spread}}
      \end{subfigure}%
      \hfill%
      \begin{subfigure}[t]{0.49\linewidth}
         \scriptsize
         \stackon{\includegraphics[width=0.325\linewidth]{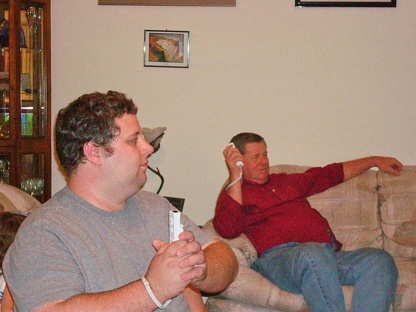}}{Image\vphan}\hfill%
         \stackon{\includegraphics[width=0.325\linewidth]{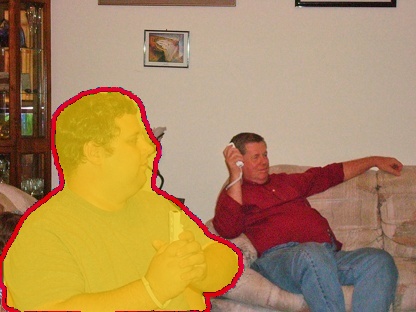}}{Prediction\vphan}\hfill%
         \stackon{\includegraphics[width=0.325\linewidth]{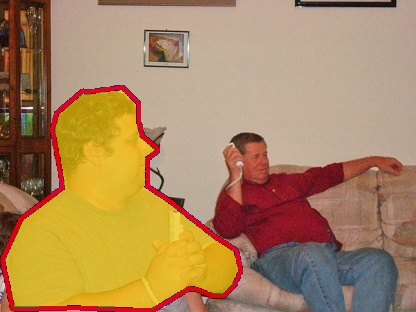}}{Ground-Truth\vphan}
         \vspace{-0.04in}
         \caption{\centering \scriptsize{a man with short hair and a grey shirt holding up a wii remote looking to the side}}
      \end{subfigure}

      \begin{subfigure}[t]{0.49\linewidth}
         \includegraphics[width=0.325\linewidth]{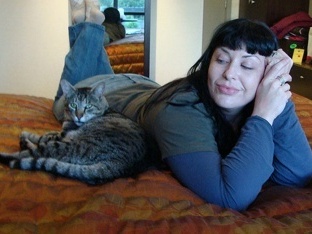}\hfill%
         \includegraphics[width=0.325\linewidth]{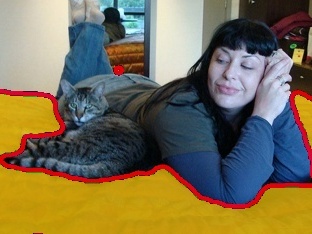}\hfill%
         \includegraphics[width=0.325\linewidth]{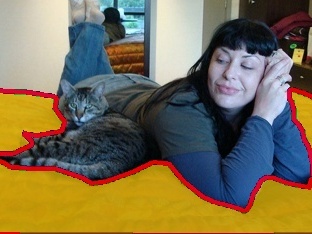}
         \vspace{-0.04in}
         \caption{\centering \scriptsize{the bed that a woman and cat are laying on}}
      \end{subfigure}%
      \hfill%
      \begin{subfigure}[t]{0.49\linewidth}
         \includegraphics[width=0.325\linewidth]{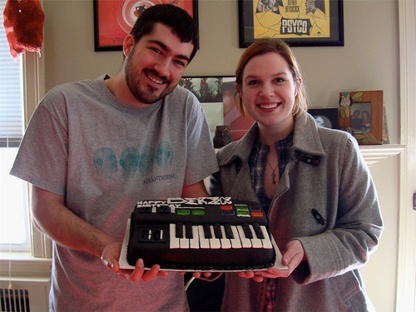}\hfill%
         \includegraphics[width=0.325\linewidth]{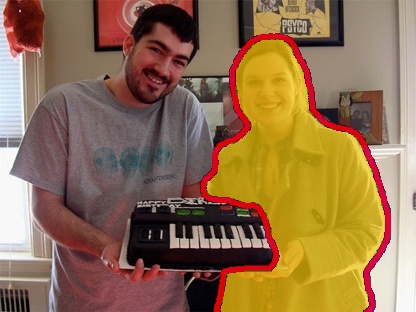}\hfill%
         \includegraphics[width=0.325\linewidth]{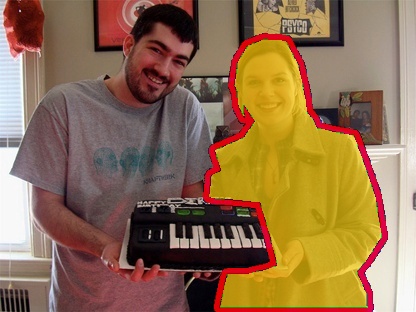}
         \vspace{-0.04in}
         \caption{\centering \scriptsize{a woman with a grey shirt holding a cake with a man}}
      \end{subfigure}%
      \caption{(Best viewed in color) Qualitative referring segmentation examples. The caption for each set of images is the input language expression.}
      \label{fig:demo}
   \end{center}
\end{figure*}
\begin{figure*}[t]
   \begin{center}
     \includegraphics[width=\linewidth]{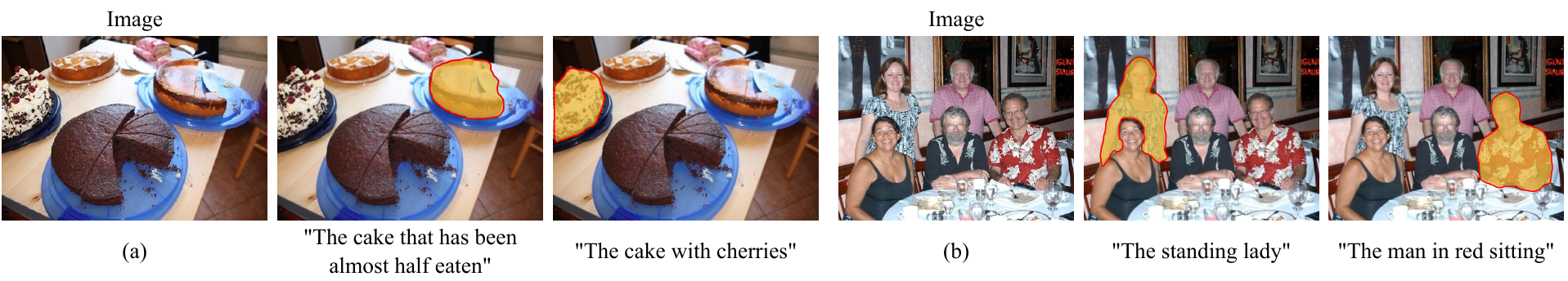}
   \end{center}
   \caption{(Best viewed in color) More examples showing our method finding different targets in a same image.}
   \label{fig:example2}
\end{figure*}

\figurename~\ref{fig:ablation} displays some example results produced by the baseline model compared with our full model. In the baseline model, we replace the {\ourdecoder} with generic transformer and remove the IMI and the LFR. The language expression of the first image is long, and the baseline model fails in comprehending this complex sentence. The second example in \figurename~\ref{fig:ablation}~\!(b) shows a more tricky case, which uses a negative sentence to target the object. The baseline model is distracted by the word ``red'' and gives wrong results, while our model successfully understands the sentence and finds the right object. From the examples, the language understanding ability of our approach is greatly enhanced compared with the baseline. This shows that the three proposed modules enable our approach to solve the hard and complex cases that the baseline model cannot handle.

\subsection{Visualizations}

In this section, we visualize some sample outputs of our model in \figurename~\ref{fig:demo}. To show the superior language understanding performance of our method, we use images and language expressions from the RefCOCOg dataset, of which language expressions are more natural and complex than other datasets. All examples in \figurename~\ref{fig:demo} have long sentences with more than 10 words, and with more than two instances appearing in the text. Example (a) has a difficult sentence and a complicated layout where three mattresses are crowded in a small room. Our model has a good context understanding of the key words ``matress'', ``pink and yellow'', ``blue'', and their relations, and does not be distracted by other mattresses and blue objects. Example (b) has a very long sentence, but most of the information is not discriminative for identifying the target, \eg, both people in the image have short hair and are looking to the side. Our model detects the informative part of the sentence and targets the right object. Example (c) shows that our model can not only identify foreground objects but is also able to detect in the backgrounds. In the language expression of example (d), three objects are mentioned: ``a woman'', ``a cake'', and ``a man''. Our model still managed to find the subject from the difficult sentence and target the instance in the image. Besides, in \figurename~\ref{fig:example2}, we show extra examples of using multiple language expressions to refer to different objects in one image. In example (a), it can be seen that our method successfully handles complex relationships and attributes such as ``has been almost half eaten'' and ``with cherry on it''. In example (b) our method can retrieval the correct object from a complex scene. The first expression tells ``standing lady'' while there are two ladies in the image. Our method found the correct one. The second expression says ``man in red sitting''. There are three information in this expression: ``man'', ``in red'', ``sitting''. From the image we can see that all of the three points are necessary to find the target, \ie, the target cannot be determined without any one of the information points. The network have to understand and combine all the information in the expression. This example shows that our network shows impressive performance on establishing the pixel-language correspondence.

\subsection{Comparison with State-of-the-Art Methods}

\begin{table*}[t]
   \renewcommand\arraystretch{1.1} 
        \centering
        \small
        \caption{Experimental results of the IoU metric. *: Google split.}\vspace{-2mm}
      \setlength{\tabcolsep}{2.3mm}{
       \begin{tabular}{l|c|c|c|c|c|c|c|c|c|c|c|c}
         \specialrule{.1em}{.05em}{.05em} 
           \multirow{2}{*}{Methods} &\multirow{2}{*}{\shortstack{Vis.\\Encoder}} & \multirow{2}{*}{\shortstack{Lang.\\Encoder}}&ReferIt&\multicolumn{3}{c|}{RefCOCO} & \multicolumn{3}{c|}{RefCOCO+} & \multicolumn{3}{c}{G-Ref} \\
       \cline{4-13}
             & & & test & val   & test A & test B & val   & test A & test B & val  & test & val*\\
           \hline
         DMN~\cite{margffoy2018dynamic} &DPN92   & SRU  & 52.81 & 49.78 & 54.83 & 45.13 & 38.88 & 44.22 & 32.29 & -     & -     & 36.76 \\
         RRN~\cite{li2018referring}     &DL-101  & LSTM & 63.63 & 55.33 & 57.26 & 53.93 & 39.75 & 42.15 & 36.11 & -     & -     & 36.45 \\
         MAttNet~\cite{yu2018mattnet}   &M-RCN   & LSTM &   -   & 56.51 & 62.37 & 51.70 & 46.67 & 52.39 & 40.08 & 47.64 & 48.61 & -     \\
         CMSA~\cite{ye2019cross}        &DL-101  & -    & 63.80 & 58.32 & 60.61 & 55.09 & 43.76 & 47.60 & 37.89 & -     & -     & 39.98 \\
         CAC~\cite{Chen_lang2seg_2019}  &R-101   & LSTM &   -   & 58.90 & 61.77 & 53.81 & -     & -     & -     & 46.37 & 46.95 & 44.32 \\
         STEP~\cite{chen2019see}        &DL-101  & LSTM & 64.13 & 60.04 & 63.46 & 57.97 & 48.19 & 52.33 & 40.41 & -     & -     & 46.40 \\
         BRINet~\cite{hu2020bi}         &DL-101  & LSTM & 63.46 & 60.98 & 62.99 & 59.21 & 48.17 & 52.32 & 42.11 & -     & -     & 48.04 \\
         CMPC~\cite{huang2020referring} &DL-101  & LSTM & 65.53 & 61.36 & 64.53 & 59.64 & 49.56 & 53.44 & 43.23 & -     & -     & 39.98 \\
         LSCM~\cite{hui2020linguistic}  &DL-101  & LSTM & 66.57 & 61.47 & 64.99 & 59.55 & 49.34 & 53.12 & 43.50 & -     & -     & 48.05 \\
         MCN~\cite{luo2020multi}        &DN-53   & GRU  &   -   & 62.44 & 64.20 & 59.71 & 50.62 & 54.99 & 44.69 & 49.22 & 49.40 & -     \\
         CMPC+~\cite{liu2021crossTPAMI} &DL-101  & LSTM & 65.58 & 62.47 & 65.08 & 60.82 & 50.25 & 54.04 & 43.47 & -     & -     & 49.89 \\
         EFN~\cite{feng2021encoder}     &R-101   & GRU  &   -   & 62.76 & 65.69 & 59.67 & 51.50 & 55.24 & 43.01 & -     & -     & 51.93 \\
         BUSNet~\cite{yang2021bottom}   &DL-101  & S-Att. &   -   & 63.27 & 66.41 & 61.39 & 51.76 & 56.87 & 44.13 & -     & -     & 50.56 \\
         CGAN~\cite{luo2020cascade}     &DL-101  & GRU  &   -   & 64.86 & 68.04 & 62.07 & 51.03 & 55.51 & 44.06 & 51.01 & 51.69 & 46.54 \\
         ISFP~\cite{liu2022instance}&DN-53  &GRU&-&{65.19} & {68.45} & {62.73} & {52.70} & {56.77} & {46.39} & {52.67} & {53.00} & 50.08 \\
         LTS~\cite{jing2021locate}      &DN-53   & GRU  &   -   & 65.43 & 67.76 & 63.08 & 54.21 & 58.32 & 48.02 & 54.40 & 54.25 & -     \\
         {VLT} \cite{dingiccv2021}    &DN-53   & GRU  &   -   & 65.65 & 68.29 & 62.73 & 55.50 & 59.20 & 49.36 & 52.99 & 56.65 & 49.76 \\
         \hline
         \textbf{Ours}$\ _\text{(CNN)}$ &{DN-53}&GRU & \textbf{69.33} & \textbf{67.88} & \textbf{70.82} & \textbf{65.02} & \textbf{56.98} & \textbf{61.26} & \textbf{50.11} & \textbf{54.79} & \textbf{58.21} & \textbf{50.96} \\
         \hline
         {ReSTR}~\cite{kim2022restr}    &ViT-B   & Transf. & 70.18 & 67.22 & 69.30 & 64.45 & 55.78 & 60.44 & 48.27 & -     & -     & 54.48 \\
         {CRIS}~\cite{wang2022cris}     &CLIP    & CLIP &   -   & 70.47 & 73.18 & 66.10 & 62.27 & 68.08 & 53.68 & 59.87 & 60.36     & - \\
         {LAVT}~\cite{yang2021lavt}     &Swin-B  & BERT &   -   & 72.73 & 75.82 & 68.79 & 62.14 & 68.38 & 55.10 & 61.24 & 62.09 & 60.50 \\
         {VLT+}~\cite{VLTPAMI}          &Swin-B  & BERT &   -   & 72.96 & 75.96 & 69.60 & 63.53 & 68.43 & 56.92 & 63.49 & 66.22 & 62.80 \\
           \hline
           \textbf{Ours}$\ _\text{(Swin)}$&{Swin-B}&BERT & \textbf{72.97} &\textbf{73.60} & \textbf{76.23} & \textbf{70.36} & \textbf{65.34} & \textbf{70.50} & \textbf{56.98} & \textbf{64.92} & \textbf{67.37} & \textbf{63.90} \\

           \specialrule{.1em}{.05em}{.05em} 
        \end{tabular}}%
        \vspace{1pt}
        \raggedright{\scriptsize{(DL: DeepLab, R: ResNet, R-MCN: ResNet+Mask R-CNN, DN: Darknet, S-Att.: Self-Attention, Transf.: Transformer)}}
        \label{tab:results}%
   \end{table*}%
\begin{table}[t]
\renewcommand\arraystretch{1.1} 
    \centering
    \small
    \caption{Results of the Precision metric on the val set of the RefCOCO.}\vspace{-2mm}
    {\begin{tabular*}{\linewidth}{@{\extracolsep{\fill}}l|ccccc}
     \specialrule{.1em}{.05em}{.05em}
     Model & {Pr@0.5} & {Pr@0.6} & {Pr@0.7} & {Pr@0.8} & {Pr@0.9}\\
     \hline\hline
     LSCM \cite{hui2020linguistic} & 70.84 & 63.82 & 53.67 & 38.69 & 12.06\\
     CMPC \cite{huang2020referring}  & 71.27 & 64.44 & 55.03 & 39.28 & 12.89\\
     MCN \cite{luo2020multi}  & 76.60 & 70.33 & 58.39 & 33.68 & 5.26\\
     LTS \cite{jing2021locate}  & 75.16 & 69.51 & 60.74 & 45.17 & 14.41 \\
     VLT \cite{dingiccv2021}  & 76.20 & - & - & - & -  \\
     \hline
     \textbf{Ours}  & \textbf{79.01}  & \textbf{74.94}  & \textbf{68.16}  & \textbf{51.21}  & \textbf{17.70} \\
     \specialrule{.1em}{.05em}{.05em}
     \end{tabular*}}%
     \vspace{-0.1in}
    \label{tab:prec}%
\end{table}%

We report the experimental results of our method on three datasets, RefCOCO~\cite{yu2016modeling}, RefCOCO+~\cite{yu2016modeling}, and RefCOCOg~\cite{mao2016generation}, to compare with previous state-of-the-art methods in \tablename~\ref{tab:results}. There are two data splitting types for the RefCOCOg dataset. One is referred to as the UMD split and the other is the Google split. The UMD split has both validation set and test set available, while the Google split only has validation set publicly available. We do experiments and report the results on both kinds of splitting. From \tablename~\ref{tab:results}, it can be seen that our method achieves superior performance on all datasets and
outperforms previous state-of-the-art methods. On RefCOCO dataset, our method is $1.5\%-2\%$ better than the previous SOTA, including VLT~\cite{dingiccv2021} and LTS~\cite{jing2021locate}. On the other two datasets, our methods also have a consistent improvement of about $1.5\%$ compared with the previous state-of-the-art methods. Besides, for a fair comparison, we also implement our model with the stronger backbone Swin-Transformer\cite{liu2021swin}. It can be seen that our model with Swin-Transformer backbone also achieves a significant improvement of around $1\%$ across most of the datasets. Especially for RefCOCO+, our model with Swin-Transformer backbone achieves about $2\%$ improvement over the previous SOTA method VLT+\cite{VLTPAMI}. This shows that our model is robust to different backbones and can achieve better performance with stronger backbones.

We also compare the Precision@X scores of the RefCOCO validation set against other methods that have data available, and the results are shown in \tablename~\ref{tab:prec}. From the Pr@0.5 row, it can be seen that our model achieves the highest score. Compared with the VLT~\cite{dingiccv2021} that also utilizes the transformer model as prediction head, our method has an over $2\%$ higher result in terms of Pr@0.5. The previous state-of-the-art method on the Pr@0.5 metric, MCN~\cite{luo2020multi}, utilizes data from both referring segmentation datasets (segmentation masks) and referring comprehension datasets (bounding boxes) in training for better locating the target, while our model only uses the segmentation mask as ground-truth. But our method achieves better targeting scores on Pr@0.5 with a large margin of 2.41\%. We attribute this to the better understanding of the language expression and the denser interaction of the information between the features from two modalities. This shows that our proposed modules leverage the information in the given language expression more effectively, and better fuse them with the vision information.

\subsection{Failure Cases}
We examine two typical categories of failure cases: (1) instances where the input expression refers to uncommon or unexpected areas. For instance, in example (a), the expression asks us to locate a ``gap between newspaper and sandwich," which, in reality, was a part of the table. Such expressions are atypical and not commonly observed in practical situations. (2) Instances where the expression is ambiguous or seeks an excessive amount of detail. In example (b), the expression ``man using oven" was used. From the picture, it was apparent that both men were operating machines in the kitchen, and both machines resembled an oven. As a result, our model highlighted both individuals. Nonetheless, if we look very carefully, the machine on top also seems like a microwave. In such cases, the expression is rather ambiguous, and the model is unable to handle them. Dealing with such situations could be an interesting topic for future research.

\begin{figure*}[t]
   \begin{center}
     \includegraphics[width=\linewidth]{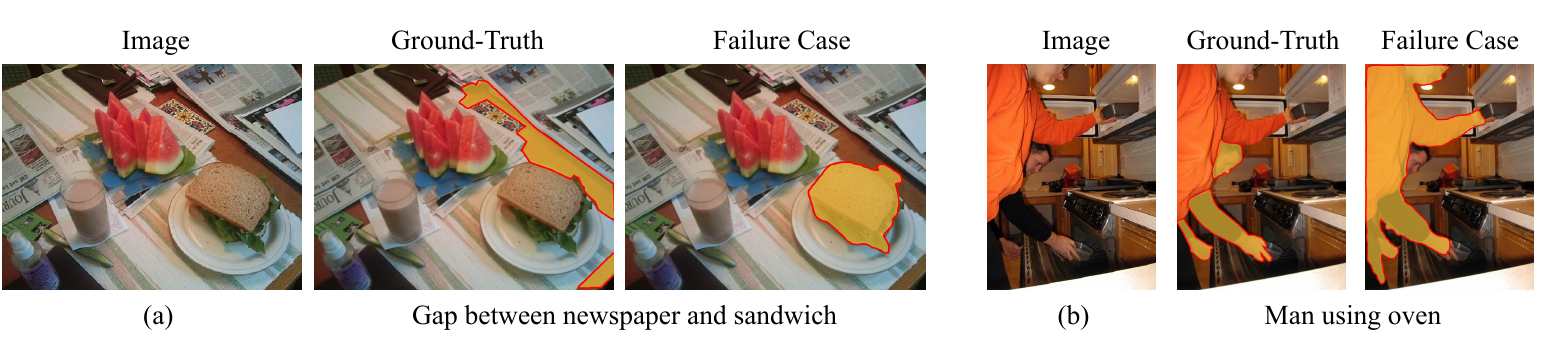}
   \end{center}
   \caption{Visualization of representative failure cases of our method.}
   \label{fig:fail_2}
\end{figure*}

\section{Conclusion}

In this work, we address the referring image segmentation problem by designing a framework that enhances the multi-modal fusion performance.
Towards this, we propose a Multi-Modal Mutual Attention ({\ourmodule}) mechanism and Multi-Modal Mutual Decoder ({\ourdecoder}) optimized for processing multi-modal information. 
Moreover, we design an Iterative Multi-Modal Interaction (IMI) scheme to further boost the feature fusing ability in the {\ourdecoder}, and introduce a Language Feature Reconstruction (LFR) module to ensure that the language information is not distorted in the network. Extensive experiments show that the proposed modules can effectively promote the interactions between the language and vision information, leading the model to achieve new state-of-the-art performance on referring image segmentation.

\bibliographystyle{IEEEtran}
\bibliography{IEEEabrv, egbib}

\vfill

\end{document}